\documentclass{article}
\usepackage{arxiv}
\usepackage[utf8]{inputenc} %
\usepackage[T1]{fontenc}    %
\usepackage[hidelinks]{hyperref}       %
\usepackage{url}            %
\usepackage{booktabs}       %
\usepackage{amsfonts}       %
\usepackage{nicefrac}       %
\usepackage{microtype}      %
\usepackage{lipsum}		%
\usepackage{graphicx}
\usepackage{threeparttable}
\usepackage{doi}
\usepackage{multirow}
\usepackage{multicol}
\usepackage{xcolor}
\usepackage{subcaption}
\usepackage[normalem]{ulem}
\usepackage{amsmath,amsfonts,bm}
\usepackage{amssymb,amsthm}
\usepackage{enumitem}
\usepackage{adjustbox}
\usepackage{blindtext}
\usepackage{cleveref}
\usepackage{float}
\usepackage{booktabs}
\usepackage{multirow}
\usepackage{array}
\usepackage{makecell} %
\usepackage[perpage]{footmisc}

\usepackage[
    backend=biber,
    style=numeric,
]{biblatex}
\newcommand{\citep}[1]{\parencite{#1}}
\setlist[itemize,1]{leftmargin=\dimexpr 18pt}
\setlist[enumerate,1]{leftmargin=\dimexpr 18pt}

\title{
\raisebox{-0.1\height}{\includegraphics[width=0.04\textwidth]{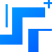}} %
Step-Audio 2 Technical Report
}
\author{\Large StepFun Audio Team}

\renewcommand{\headeright}{Technical Report}
\renewcommand{\undertitle}{}
\renewcommand{\shorttitle}{\raisebox{-0.12\height}{\includegraphics[width=0.02\textwidth]{figures/title-logo.png}}
Step-Audio 2}

\begin{document}
\large

\maketitle
\begin{abstract}
This paper presents Step-Audio 2, an end-to-end multi-modal large language model designed for industry-strength audio understanding and speech conversation. By integrating a latent audio encoder and reasoning-centric reinforcement learning (RL), Step-Audio 2 achieves promising performance in automatic speech recognition (ASR) and audio understanding. To facilitate genuine end-to-end speech conversation, Step-Audio 2 incorporates the generation of discrete audio tokens into language modeling, significantly enhancing its responsiveness to paralinguistic information such as speaking styles and emotions. To effectively leverage the rich textual and acoustic knowledge in real-world data, Step-Audio 2 integrates retrieval-augmented generation (RAG) and is able to call external tools such as web search to mitigate hallucination and audio search to switch timbres. Trained on millions of hours of speech and audio data, Step-Audio 2 delivers intelligence and expressiveness across diverse conversational scenarios. Evaluation results demonstrate that Step-Audio 2 achieves state-of-the-art performance on various audio understanding and conversational benchmarks compared to other open-source and commercial solutions. Please visit \url{https://github.com/stepfun-ai/Step-Audio2} for more information.

\end{abstract}

\begin{figure}[htb!]
\centering
\includegraphics[width=0.8\textwidth]{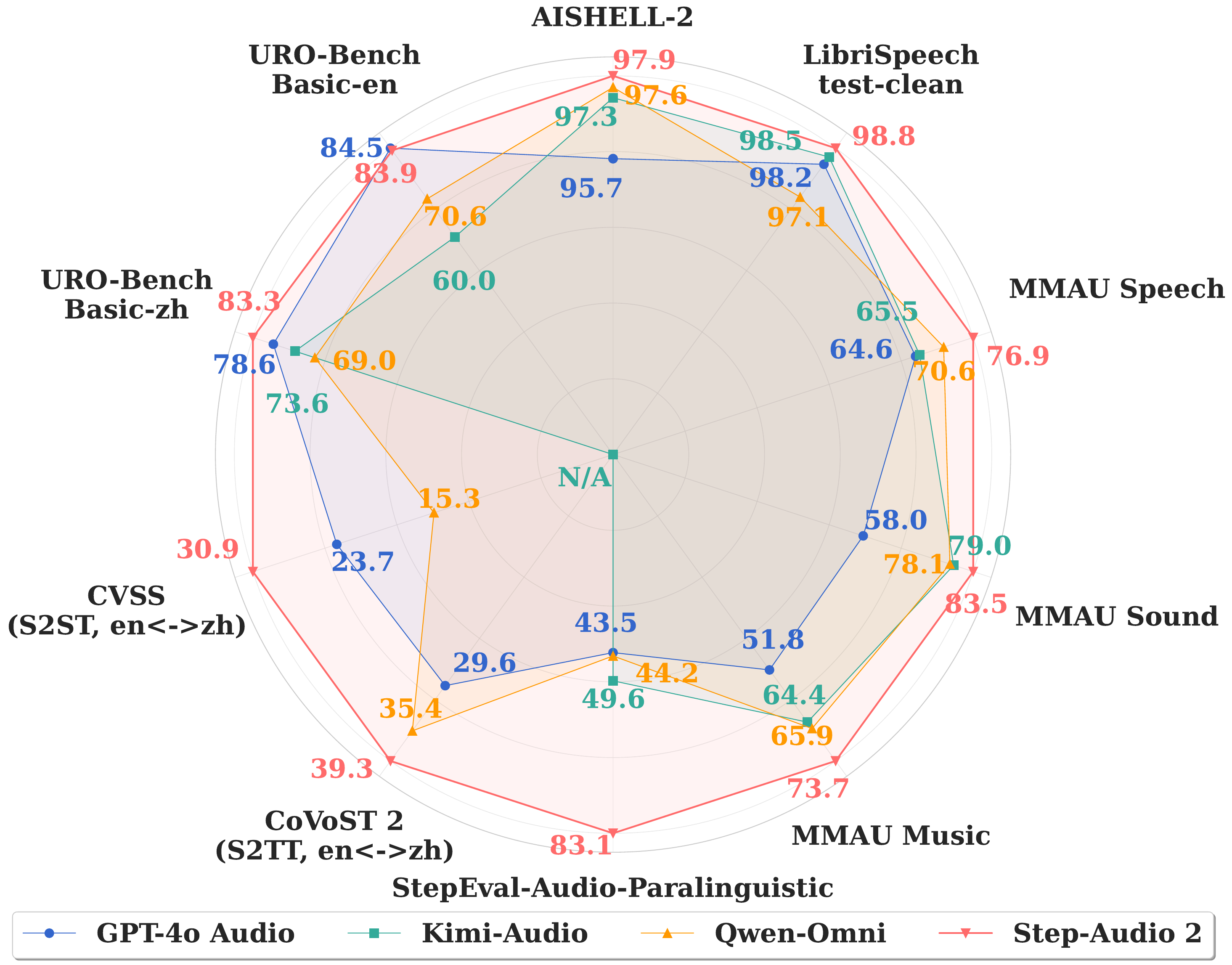}
\caption{Performance comparision of GPT-4o Audio$^1$, Kimi-Audio$^2$, Qwen-Omni$^3$ and Step-Audio 2 and on various benchmarks.}
\label{fig:radar}
\end{figure}
\section{Introduction}

With the rapid development of large language models and audio processing technology,
large audio language models (LALMs) have demonstrated their superiority 
over conventional approaches in 
various speech and audio processing tasks.
GPT-4o is first introduced and is pioneering the development of end-to-end speech interaction without intermediate textual conversions.
Subsequently, many open-sourced LALMs \cite{chen2025minmo,chu2024qwen2,defossez2024moshi,ding2025kimi,fang2024llama,huang2025stepaudioaqaa,huang2025step,nguyen2024spiritlminterleavedspoken,xie2024mini,xie2024mini2,xu2025qwen25omnitechnicalreport,zeng2024glm} are emerged, advancing multi-modal large language model capabilities in various speech and audio domains.
Among these approaches, 
Qwen-Audio~\cite{chu2023qwen} and Qwen2-Audio~\cite{chu2024qwen2} perform audio analysis and generate textual responses to speech instructions. Qwen2.5-Omni~\cite{xu2025qwen25omnitechnicalreport} implements a thinker-talker architecture to enable full-duplex I/O during speech conversations.
More recently, Kimi-Audio~\cite{ding2025kimi} has achieved impressive results on multiple speech and audio understanding benchmarks. In parallel, we have introduced Step-Audio~\cite{huang2025step} and Step-Audio-AQAA~\cite{huang2025stepaudioaqaa}, the first LALMs to unify speech understanding and generation through discrete audio tokens at a scale of 130 billion parameters.

However, %
existing LALMs still face challenges
in achieving natural and intelligent speech interaction.
Previous LALMs such as Spirit LM~\cite{nguyen2024spiritlminterleavedspoken} and GLM-4-Voice~\cite{zeng2024glm} mainly focus on aligning the semantic information in speech inputs to text modal,
neglecting the para-linguistic information which is also crucial for intentional understanding.
Although LALMs including Qwen-Audio~\cite{chu2023qwen}, Qwen2-Audio~\cite{chu2024qwen2} and Audio Flamingo series \cite{ghosh2025audioflamingo2, goel2025audioflamingo3,kong2024audioflamingonovelaudio} are capable of comprehending such information, they typically generate only textual outputs and fail to further utilize this capability to produce coherent and expressive responses in speech conversations.
Moreover, due to the complexities of multi-modal modeling, existing LALMs frequently suffer from hallucination and offer limited choices of timbres and speaking styles~\cite{defossez2024moshi,ding2025kimi}, lacking access to real-world textual and acoustic knowledge.

\footnotetext[1]{GPT-4o Audio is evaluated with gpt-4o-transcribe for ASR and gpt-4o-audio-preview-2025-06-03 for others via official API.}
\footnotetext[2]{Kimi-Audio is excluded from translation evaluations since it consistently ignores prompts.}
\footnotetext[3]{Qwen-Omni is evaluated with Qwen2.5-Omni for MMAU and speech-to-text translation, and qwen-omni-turbo-2025-03-26 for others via official API.}

To address these issues and step into the next generation of multi-modal large language models,
we present \textbf{Step-Audio 2}, an end-to-end large audio language model
with industry-strength audio perception and speech interaction. %
Step-Audio 2 directly processes raw audio as input and outputs discrete text and audio tokens and has fewer parameters than Step-Audio~\cite{huang2025step}. 
Beyond capturing semantic information in speech, the model also comprehends para-linguistic and non-vocal information in audio.
By leveraging chain-of-thought (CoT) reasoning and reinforcement learning (RL), 
Step-Audio 2 further utilizes such multi-modal information to generate expressive speech responses coherent to 
different conversation scenarios.
To ground the model with real-world knowledge,
Step-Audio 2 incorporates retrieval-augmented generation (RAG) and the capability to utilize various external tools, including web search and audio search, to provide more reliable and expressive responses.
Specifically, we present an audio search as a tool unique to LALMs,
enabling seamless speech retrieval via voice instructions and allowing the model to switch timbres and speaking styles based on the retrieved speech.

\begin{figure}[tb]
\centering
\includegraphics[clip, trim=0cm 0.2cm 0cm 0cm, width=\textwidth]{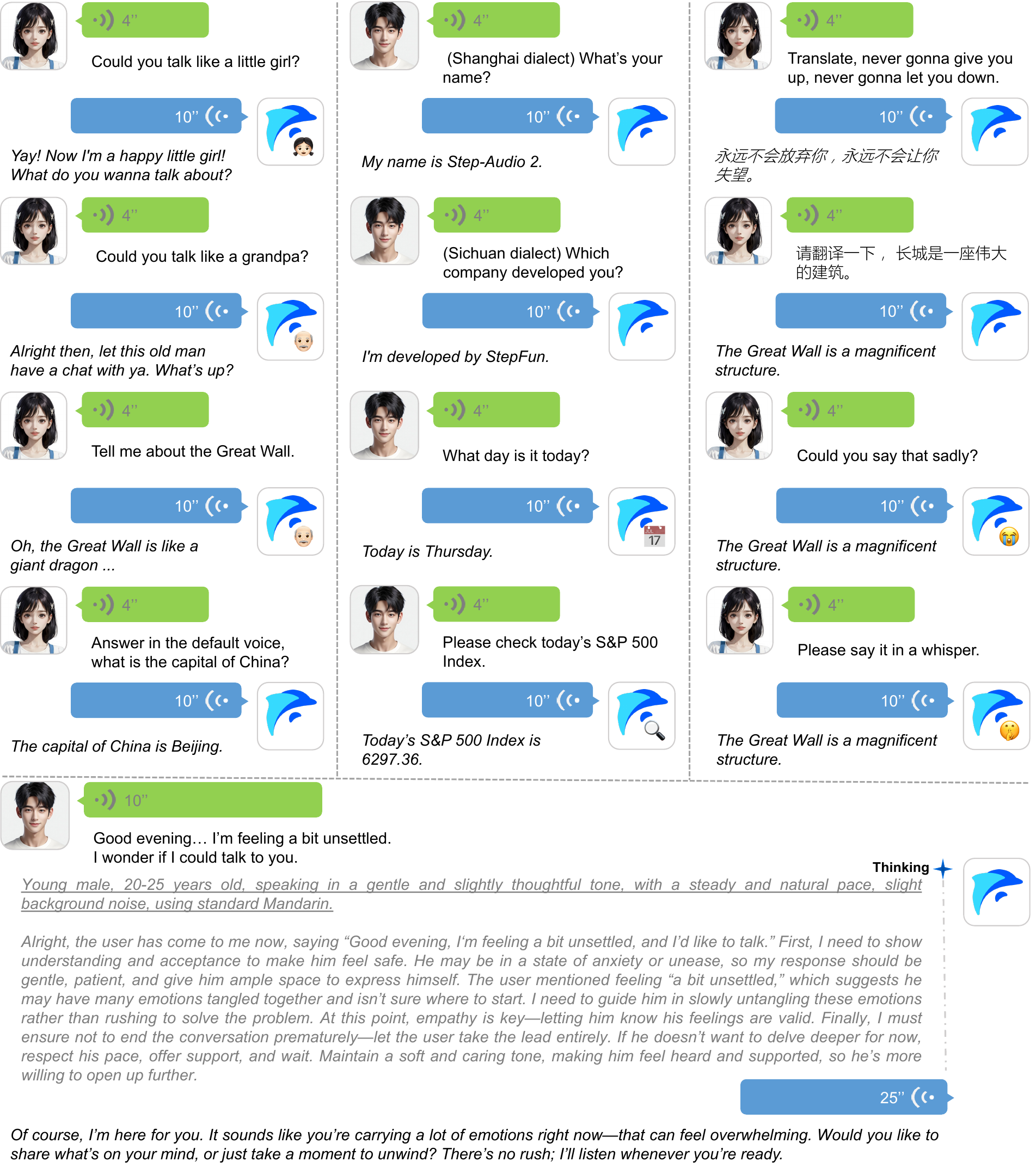}
\caption{Illustration of the applications of Step-Audio 2 across various speech conversation scenarios.}
\label{fig:usage}
\end{figure}

To ensure its intelligence and
expressiveness in diverse conversational scenarios, 
we carefully design a multi-stage training strategy to train Step-Audio 2 on 680 billion tokens of text data and 8 million hours of real and synthesized audio data.
Evaluation results shown in Figure \ref{fig:radar} demonstrate that Step-Audio~2 achieves state-of-the-art performance in a series of audio tasks, including automatic speech recognition (ASR) on multiple languages, audio understanding, 
speech-to-speech translation and speech-to-speech conversation.
Typical usages of Step-Audio 2 are also illustrated in Figure \ref{fig:usage}.

\section{Related Work}

\subsection{Speech and audio understanding}

Recent advances in large language models (LLMs)~\cite{bai2023qwentechnicalreport,grattafiori2024llama3herdmodels,openai2023gpt4,openai2022chatgpt} have extended their application to a wide range of speech and audio understanding tasks, such as audio captioning, sound event detection, automatic speech recognition, audio classification, and audio-driven creative generation.
A prevalent approach~\cite{chu2024qwen2,deshmukh2023pengi,gong2023joint,gong2023listen,ParalinGPT,tang2023salmonn} involves pairing speech encoders with lightweight, trainable adapters that project audio features into a textual embedding space compatible with LLMs.
Building on this foundation, recent studies have further explored how to incorporate paralinguistic information such as emotion, intonation and speaker style, enabling LLMs to move beyond pure linguistic comprehension. For instance, ParalinGPT~\cite{ParalinGPT} focuses on enhancing a powerful text-based language model by integrating continuous speech embeddings, enabling it to capture paralinguistic signals such as emotion and prosody. SALMONN~\cite{tang2023salmonn} adopts a multi-modal strategy by freezing speech encoders Whisper~\cite{radford2023robust} and BEATs~\cite{BEATs}, and connecting their outputs to an LLM via a window-level Q-Former, enabling joint modeling of linguistic and acoustic features. Seed-ASR~\cite{Seedasr} integrates LUISE-based speech representations with instructions and context, using context-aware SFT to capture semantic information. AudioPaLM~\cite{rubenstein2023audiopalm} combines PaLM-2~\cite{anil2023palm2technicalreport} and AudioLM~\cite{borsos2023audiolm}, unifying linguistic knowledge with paralinguistic features like speaker identity and intonation. LLM-based approaches~\cite{chu2023qwen,chu2024qwen2} increasingly rely on pretrained audio encoders such as Wav2Vec~\cite{wav2vec2}, HuBERT~\cite{HuBERT}, Whisper~\cite{radford2023robust}, and WavLM~\cite{WavLM} to extract rich semantic representations from speech. At the same time, the extensive text knowledge and contextual reasoning capabilities stored in LLMs can provide valuable semantic guidance for understanding tasks.

\subsection{Text-to-speech synthesis}

Text-to-Speech (TTS) technology has made remarkable strides in recent years, evolving from traditional concatenative and statistical parametric approaches~\cite{deepvoice3,fastspeech2,tacotron2,wang2017tacotron} to codec-based TTS systems.
Codec language models leverage a speech codec to extract discrete representations of speech~\cite{encodc,defossez2024moshi,ji2024wavtokenizer,snac,,xin2024bigcodec,soundstream,zhang2025distinctive,zhang2023speechtokenizer} and utilize either autoregressive~\cite{ding2025kimi,zhang2023speechgpt} or masked language models~\cite{wang2024maskgct} to predict the corresponding speech tokens. These tokens are then synthesized into waveforms using codec vocoders. VALL-E~\cite{wang2023neural} marked a significant breakthrough in this area. It uses an autoregressive model to generate coarse codec codes, followed by a non-autoregressive model for the fine codes. Unlike VALL-E, which predicts acoustic tokens from phonemes and requires transcripts, SPEAR-TTS~\cite{SPEARTTS} uses a two-stage architecture with self-supervised audio prompts to clone unseen voices from just 3 seconds of speech. SparkTTS~\cite{wang2025spark} introduces BiCodec, a single-stream speech codec that encodes linguistic content as compact semantic tokens and speaker characteristics as fixed-length global tokens. Instead of relying on non-autoregressive models to predict residual discrete codes, methods like TorToiseTTS~\cite{TorToiseTTS}, CosyVoice~\cite{du2024CosyVoice}, CosyVoice~2~\cite{du2024cosyvoice2}, MiniMax-Speech~\cite{zhang2025minimaxspeech} and SEED-TTS~\cite{seedtts} adopt diffusion or flow-matching techniques as a second stage to reconstruct mel-spectrograms or continuous representations enriched with fine-grained acoustic and semantic details. Recent work, Kimi-Audio~\cite{ding2025kimi} combines Whisper features and semantic tokens for efficient modeling, with dual heads and a flow-matching detokenizer plus BigVGAN~\cite{lee2023bigvgan} for low-latency, expressive synthesis.

\subsection{Speech-to-speech translation}
Speech-to-speech translation (S2ST) is a crucial technology for eliminating communication barriers across languages. Traditional S2ST systems~\cite{wahlster2013verbmobil,wu2016google} typically adopt a cascaded pipeline consisting of automatic speech recognition (ASR), machine translation (MT), and TTS modules. Earlier studies~\cite{jia2019direct,jia2022translatotron,le2024transvip,lee2021textless} have shifted toward direct approaches that bypass intermediate textual representations, aiming for lower latency and better preservation of prosody and speaker characteristics. Two main types of direct S2ST methods have emerged, which are known as speech-to-spectrogram translation and speech-to-unit translation. Both directly generate target speech representations from the source speech without relying on textual transcriptions. A representative of the former is Translatotron~\cite{jia2019direct}, the first end-to-end model to translate source speech directly into target spectrograms. Translatotron 2~\cite{jia2022translatotron}, further improves translation quality through a two-pass decoding mechanism~\cite{lee2021textless}. In contrast, speech-to-unit models predict discrete acoustic tokens rather than spectrograms, which are typically extracted using self-supervised speech encoders such as HuBERT~\cite{HuBERT} or WavLM~\cite{WavLM}. For instance, TransVIP~\cite{le2024transvip} employs a joint encoder-decoder architecture that first generates target text and residual vector quantization (RVQ) codes in the initial layer, followed by a non-causal language model that refines RVQ predictions in subsequent layers.

\subsection{Speech-to-text and speech-to-speech conversation}

Based on whether the LLM can directly understand and generate speech representations, existing systems can be categorized into end-to-end large audio language models and cascaded large audio language models. The former directly models audio inputs and outputs within a unified framework, while the latter relies on a modular pipeline involving separate ASR, LLM, and TTS components. Traditional speech-to-text and speech-to-speech systems typically adopt a cascaded architecture, as exemplified by AudioGPT~\cite{huang2024audiogpt} and Spoken-LLM~\cite{lin2024advancing}. However, the ASR + LLM + TTS pipeline incurs high latency and modular mismatches. This has spurred interest in unified end-to-end architectures for faster and more seamless integration. A major milestone in this direction is GPT-4o~\cite{hurst2024gpt}, which supports direct end-to-end speech interaction without requiring intermediate textual conversions. Recently, several new end-to-end LALMs~\cite{defossez2024moshi,fang2024llama,xie2024mini,xie2024mini2,xu2025qwen25omnitechnicalreport} for speech-to-speech conversation have emerged. For instance, Moshi~\cite{defossez2024moshi} improves efficiency with an RQ-Transformer that generates text and audio tokens simultaneously. Similarly, Mini-Omni~\cite{xie2024mini} generates speech and text responses in parallel, following a strategy similar to MusicGen~\cite{musicgen}, which enables lower first-token latency compared to interleaved generation designs. LUCY~\cite{gao2025lucylinguisticunderstandingcontrol} builds on the Mini-Omni architecture with enhancements for emotional expressiveness, naturalness, and informativeness in speech generation. It utilizes curated synthetic data and optimizes the training and decoding pipelines to handle multi-turn dialogue and function-call scenarios. Mini-Omni2~\cite{xie2024mini} further extends Mini-Omni framework by integrating multimodal understanding and full-duplex interaction capabilities. LLaMA-Omni~\cite{fang2024llama} introduces a streaming, non-autoregressive speech decoder based on Connectionist Temporal Classification, enabling direct and efficient generation of discrete audio tokens without relying on step-by-step prediction. Freeze-Omni~\cite{wang2024freeze}, on the other hand, freezes the LLM parameters during training, preserving its original capabilities while achieving low-latency speech-to-speech interaction through streaming and decoder integration. Qwen2.5-Omni~\cite{xu2025qwen25omnitechnicalreport} supports multimodal input and simultaneous text-speech output via a thinker-talker architecture, using TMRoPE for improved audio-visual synchronization through explicit temporal encoding.

\begin{figure}[t]
\centering
\includegraphics[clip, trim=0.75cm 0.3cm 0.3cm 0cm, width=\textwidth]{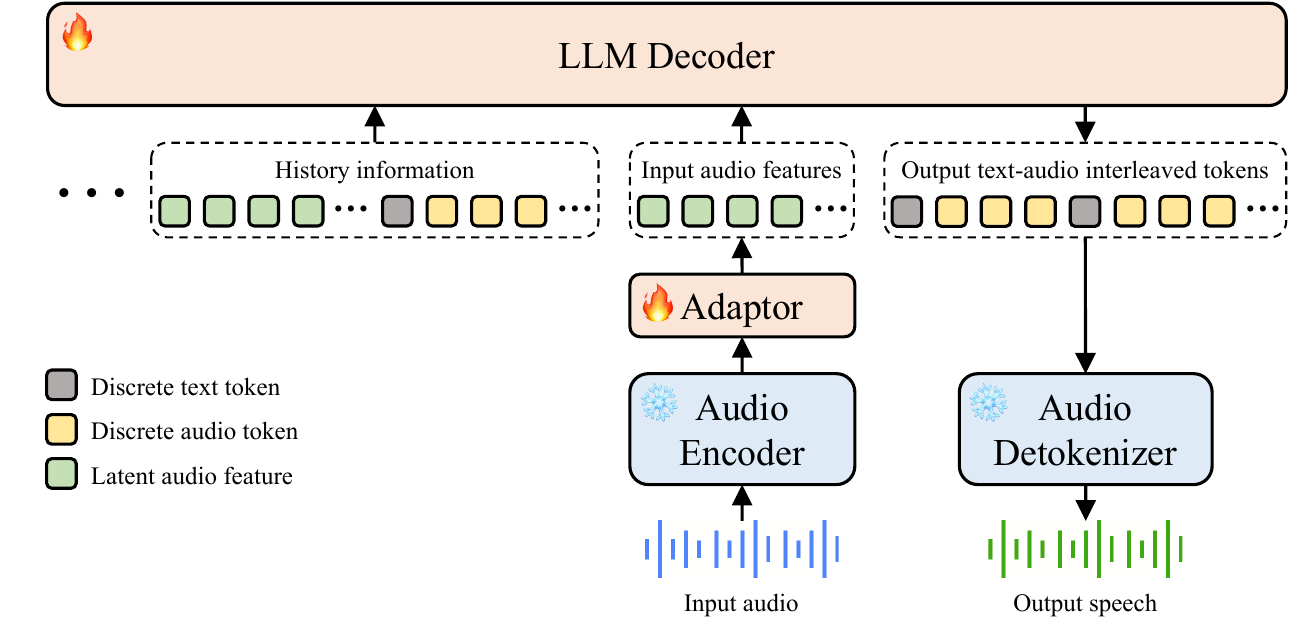} 
\caption{Architecture of the Step-Audio 2.}
\label{fig:architecture}
\end{figure}

\section{Methodology}
\subsection{Architecture}

Different from our previous Step-Audio~\cite{huang2025step},
Step-Audio 2 further integrates the generation of audio tokens into language modeling,
achieves end-to-end audio perception and generation.
As shown in Figure~\ref{fig:architecture}, Step-Audio 2 consists of
an audio encoder,
an audio adaptor,
an LLM decoder 
and an audio detokenizer.

The audio encoder is pretrained on various speech and audio understanding tasks including ASR, speaker age and gender prediction, audio event detection, etc.
The audio encoder has an output frame rate of 25 Hz and is frozen during the entire training process. 
An audio adaptor with a downsampling rate of 2 is employed to connect the audio encoder to LLM, thereby reducing the output frame rate of the audio encoder to 12.5 Hz. 

The LLM decoder directly takes the latent audio features from the audio adaptor as input,
and outputs an interleaved sequence of discrete text and audio tokens.
We employ the tokenizer from CosyVoice 2 \cite{du2024cosyvoice2} as the audio tokenizer.
And the text and audio tokens are interleaved \cite{huang2025stepaudioaqaa,huang2025step,zeng2024glm} at a fixed ratio and padded at the end to meet the ratio. %
The audio tokens are then extracted from the interleaved sequence and consumed by the audio detokenizer to generate the output waveform.
The input audio features and output interleaved sequences 
are then pre-filled as the history information for the next round of conversation.

To provide more accurate responses and expand interactive capabilities, %
we 
design tools to retrieve audio, current date and time, weather forecast and web content directly with explicit or implicit voice inputs.
Notably, 
we propose the audio search tool,
a novel tool with a voice library of hundreds of thousands of speeches with their corresponding transcriptions and descriptions.
With the retrieved speech from audio search, Step-Audio 2 is able to mimic the speaking style
or switching timbre according to the speech.
During inference,
the retrieved information is appended after the input audio features
before generating speech outputs.

Similar to Step-Audio~\cite{huang2025step} and Step-Audio-AQAA~\cite{huang2025stepaudioaqaa}
, Step-Audio 2’s audio detokenizer also consists of a Flow Matching module and a HiFi-GAN~\cite{kong2020hifigangenerativeadversarialnetworks} vocoder. The Flow-Matching module generates Mel spectrograms from the output audio tokens, while the vocoder further converts the Mel spectrograms into waveforms.
For Flow-Matching, we incorporate a CNN-based encoder layer after each self-attention module within the transformer block and train the model on 200,000 hours of high-quality speech. This enhancement significantly improves its Mel spectrogram reconstruction capability, leading to substantial gains in both pronunciation accuracy and timbre similarity.

Step-Audio 2 employs the same deployment infrastructure used in Step-Audio~\cite{huang2025step} and Step-Audio-AQAA~\cite{huang2025stepaudioaqaa}, which includes a voice activity detection (VAD) module to filter out input speeches and achieves real-time voice conversation. %

\subsection{Pre-training}

Step-Audio 2 model is initialized with a textual LLM and %
then continually pre-trained on 1.356T tokens of textual and audio data
over 21 days.

We first utilize
100B tokens
of ASR data to facilitate effective alignment between speech and text feature spaces within the adaptor.
During this phase, both the audio encoder and LLM are frozen, with only the adaptor being trained.
We conduct training for 12K steps at an 8,192 sequence length.
And the learning rate decays from $10^{-4}$ to $2\times10^{-5}$. 

We then extend the tokenizer of the textual LLM with 6.6K audio tokens.
To properly embed the new audio tokens and preserve the model's textual capabilities,
the model is then trained on 128B tokens of text data and 128B tokens of audio data.
Specifically, audio data includes 80B, 32B and 16B tokens of TTS, speech-to-speech conversation and utterance-level text-speech interleaved continuation data respectively.
The sequence length is increased to 16,384.
And the learning rates of the LLM,
adaptor, embedding layer
and output layer
are set to $2\times10^{-5}$, 
$5\times10^{-5}$, 
$5\times10^{-5}$, 
and $4\times10^{-5}$ respectively.

We then introduce our main pre-training process and further train the model on another 800B tokens of text and audio data.
We unify the learning rates to $2\times10^{-5}$ 
and
employ
400B tokens of textual data and 
42B, 120B, 8B, 30B, 5B, 45B and 150B tokens of ASR, TTS, speech-to-text translation, text-to-speech translation, speech-to-text continuation, utterance-level text-speech interleaved continuation and speech-to-speech conversation data respectively.

We finally employ 200B tokens of high-quality text and audio data to introduce a wider array of tasks and cooldown the model.
We employ
24.6B,
12.4B,
2.4B,
and 3.6B
tokens of audio data 
for 
multilingual and dialectal ASR, 
TTS, 
paralinguistic information understanding, 
speech-to-text translation respectively.
Besides, 
we develop a conversational speech synthesis pipeline to synthesize 6B, 15B and 36B tokens of audio data for speech-to-speech translation, utterance-level text-speech interleaved conversation and speech-to-speech conversation.
To ensure the vocal diversity in the synthesized speech, the system references a library of approximately 50k unique speakers.
We balance the audio data with 100B tokens of high-quality text data and
the learning rate decays from $2\times10^{-5}$ to $5\times10^{-6}$.

After this comprehensive pre-training procedure, the model 
has acquired strong audio understanding and generation capabilities
while maintaining its textual performance inherited from the initial textual LLM.

\subsection{Supervised fine-tuning}

We subsequently perform a large-scale, multi-task supervised fine-tuning (SFT) procedure \cite{wei2021finetuned} to
instruct the model
to follow human intention in fluid conversations
and master core tasks.
We select audio data 
from open-source and proprietary data to ensure broad coverage and high quality.
The model is trained on 4B tokens of text and audio data for a single epoch.
And the learning rate decays from $10^{-5}$ to $10^{-6}$.

Specifically, we leverage extensive corpora such as 
GigaSpeech~\cite{gigaspeech}, 
WenetSpeech~\cite{zhang2022wenetspeech10000hoursmultidomain}, 
and other in-house data to enhance the model's performance in multilingual and multi-dialect ASR scenarios.
We reformat existing datasets for audio event classification and audio captioning, such as AudioSet~\cite{7952261} and AudioCaps~\cite{kim2019audiocaps}, into speech question-answer pairs for audio understanding.
To capture paralinguistic information beyond just semantics, we introduce a detailed speech captioning task and build an in-house dataset, requiring the model to generate comprehensive textual descriptions encompassing 11 paralinguistic and environmental aspects.

We employ high-quality, professionally labeled data collected in-house for TTS. We utilize the Chinese-to-English and English-to-Chinese subsets from the CoVoST 2~\cite{wang2020covost2massivelymultilingual} dataset for speech-to-speech translation.

We leverage high-quality in-house textual data for classic text-to-text conversation.
Multiple LLMs are then employed to rewrite these text conversations as dialogue scripts with a more natural, colloquial style. We randomly insert emotion and speed instructions into the generated scripts to enable basic emotion and speaking style control. 
The scripts are then synthesized into speech conversations using our conversation synthesis pipeline.

We construct approximately 1K dialogue scripts in text for each type of external tools.
Within these scripts, instructions with explicit or implicit tool invocation intentions and their corresponding statements are inserted into common dialogues. The scripts are then synthesized into speech conversations using our conversation synthesis pipeline.

Besides, we construct and employ two reasoning-centric datasets during SFT to cold-start the subsequent reinforcement learning process.
First, we build a dataset to enable and robust audio understanding in complex acoustic scenarios,
by combining multiple audios from AudioSet and AudioCaps, thereby creating intricate acoustic environments.
To better address and respond to the paralinguistic information in speech conversations,
we synthesize a speech conversation dataset with our conversation synthesis pipeline,
based on dialogue scripts with appropriate emotion descriptions generated from textual LLMs.
Subsequently, a textual LLM with reasoning capabilities is employed to produce question–answer pairs with explicit step-by-step reasoning traces, according to the audio mixing recipes or the generated dialogue scripts.

\subsection{Reinforcement learning}

To enhance the model's reasoning capabilities in audio understanding and speech interaction,
we implement a multi-stage reinforcement learning strategy.
We leverage our reasoning-centric datasets from SFT and utilize two stages of proximal policy optimization (PPO)~\cite{rafailov2023direct} to optimize reasoning efficiency for real-time audio engagement.
In the first stage, a binary reward function is employed to limit the thinking sequence length to a predefined maximum.
This reward function assigns a value of 1 for reasoning that is appropriately concise (neither empty nor excessively long) and 0 otherwise.
Training is conducted for 60 iterations with a global batch size of 64, using an actor learning rate of $1 \times 10^{-6}$ and a critic learning rate of $2.5 \times 10^{-6}$.
The second stage transitions from binary rewards to learned preference scoring, utilizing a trained reward model to evaluate response quality.
This stage involves an additional 120 iterations while maintaining the same batch size and learning rate settings. Finally, we incorporate group relative policy optimization (GRPO)~\cite{rafailov2023direct} for 400 iterations to further improve the model's audio perceptual abilities.

\section{Evaluation}

\subsection{Automatic speech recognition}
As the most critical component of audio understanding and speech interaction, we first evaluate the model's capability in automatic speech recognition. We evaluate Step-Audio 2 across six Chinese test sets, four English test sets, three multilingual test sets (Japanese, Cantonese, Arabic), and six in-house Chinese dialect and accented Mandarin test sets.
For comparative analysis, we utilize top-performing models from both open-source and commercial domains as baselines, including Doubao LLM ASR\footnote{Doubao LLM ASR refers to~\url{https://www.volcengine.com/docs/6561/1354868}}, GPT-4o Transcribe\footnote{GPT-4o Transcribe is evaluated using its latest model, gpt-4o-transcribe, via its official API.}, Kimi-Audio~\cite{ding2025kimi}, and Qwen-Omni.
We prefer GPT-4o Transcribe than GPT-4o Audio since the formers provide stronger results.
Notably, Doubao LLM ASR and GPT-4o Transcribe represent specialized ASR systems that achieve leading-edge performance.

We evaluate all the models without specifying language\footnote{We evaluate without specifying language to ensure a fair comparison. Notably, Qwen-Omni lacks a language-independent testing approach, specifying language may yield better results.} and summarize the results in Table \ref{tab:asr}.
Step-Audio 2 outperforms existing open-source and commercial ASR models in both general English and Chinese recognition,
achieving an average word error rate (WER) of 3.14\% on English 
and an average character error rate (CER) of 3.08\% on Chinese test sets.
Moreover, Step-Audio 2 offers comparable results to GPT-4o Transcribe on Arabian and Japanese recognition, to Qwen-Omni on Cantonese recognition, demonstrating its capability in multilingual speech recognition.
In addition, Step-Audio 2 achieves the lowest average CER among 4 in-house Chinese accented Mandarin and 2 dialect test sets.
These results highlight the superiority of Step-Audio 2 in understanding the semantic information in speech.

\begin{table}[htb]
\centering
\caption{Comparison between Doubao LLM ASR, GPT-4o Transcribe, Kimi-Audio, Qwen-Omni and Step-Audio 2, on character (for Chinese, Cantonese and Japanese) and word (for Arabian and English) error rates among multiple ASR test sets. N/A indicates that the language is not supported.}
\label{tab:asr}
\begin{tabular}{lcccccc}
\toprule
\textbf{Category} & \textbf{Test set} & \textbf{\makecell{Doubao\\LLM ASR}} & \textbf{\makecell{GPT-4o \\Transcribe}} & \textbf{\makecell{Kimi-\\Audio}} & \textbf{\makecell{Qwen-\\Omni}} & \textbf{\makecell{Step-\\Audio 2}} \\
\midrule
\multirow{5}{*}{\textbf{English}} 
& Common Voice & 9.20 & 9.30 & 7.83 & 8.33 & \textbf{5.95} \\
& FLEURS English & 7.22 & \textbf{2.71} & 4.47 & 5.05 & 3.03 \\
& LibriSpeech clean & 2.92 & 1.75 & 1.49 & 2.93 & \textbf{1.17} \\
& LibriSpeech other & 5.32 & 4.23 & 2.91 & 5.07 & \textbf{2.42} \\
\cmidrule(lr){2-7}
& \textbf{Average} & 6.17 & 4.50 & 4.18 & 5.35 & \textbf{3.14} \\
\cmidrule(r){1-7}
\multirow{7}{*}{\textbf{Chinese}}
& AISHELL & 0.98 & 3.52 & \textbf{0.64} & 1.17 & 0.63 \\
& AISHELL-2 & 3.10 & 4.26 & 2.67 & 2.40 & \textbf{2.10} \\
& FLEURS Chinese & 2.92 & \textbf{2.62} & 2.91 & 7.01 & 2.68 \\
& KeSpeech phase1& 6.48 & 26.80 & 5.11 & 6.45 & \textbf{3.63} \\
& WenetSpeech meeting & 4.90 & 31.40 & 5.21 & 6.61 & \textbf{4.75} \\
& WenetSpeech net & \textbf{4.46} & 15.71 & 5.93 & 5.24 & 4.67 \\
\cmidrule(lr){2-7}
& \textbf{Average} & 3.81 & 14.05 & 3.75 & 4.81 & \textbf{3.08} \\
\cmidrule(r){1-7}
& FLEURS Arabian & N/A & \textbf{11.72} & N/A & 25.13 & 14.22 \\
\textbf{Multilingual}& Common Voice yue & 9.20 & 11.10 & 38.90 & \textbf{7.89} & 7.90 \\
& FLEURS Japanese & N/A & \textbf{3.27} & N/A & 10.49 & 3.18 \\
\cmidrule(r){1-7}
 & Anhui accent & \textbf{8.83} & 50.55 & 22.17 & 18.73 & 10.61 \\
 & Guangdong accent & 4.99 & 7.83 & \textbf{3.76} & 4.03 & 3.81 \\
 & Guangxi accent & 3.37 & 7.09 & 4.29 & \textbf{3.35} & 4.11 \\
\textbf{In-house} & Shanxi accent & 20.26 & 55.03 & 34.71 & 25.95 & \textbf{12.44} \\
 & Sichuan dialect & \textbf{3.01} & 32.85 & 5.26 & 5.61 & 4.35 \\
& Shanghai dialect & 47.49 & 89.58 & 82.90 & 58.74 & \textbf{17.77} \\
\cmidrule(lr){2-7}
& \textbf{Average} & 14.66 & 40.49 & 25.52 & 19.40 & \textbf{8.85} \\
\bottomrule
\end{tabular}
\end{table}

\subsection{Paralinguistic information understanding}

We then evaluate how Step-Audio 2 understands the paralinguistic information in speech beyond the semantic information.
To this end, we introduce StepEval-Audio-Paralinguistic, a speech-to-speech benchmark that evaluates the model's understanding of paralinguistic information across 11 dimensions using single-turn question answering.

StepEval-Audio-Paralinguistic comprises 550 speech samples evenly distributed across 11 tasks. We initially collect 400 Chinese speech clips for 8 of these tasks from public podcast recordings, encompassing gender, age, timbre, emotion, pitch, rhythm, speaking speed, speaking style, and vocal activity prediction or description. For sound event, scenario, and vocal sound detection or description, we source 50 event-related, 50 environmental, and 50 vocal sounds from AudioSet~\cite{7952261}, CochlScene~\cite{9979822}, and VocalSound~\cite{9746828}, respectively. All original recordings are shorter than 30 seconds and uniformly resampled to 24,000 Hz, with annotations provided by professional groups in open-set natural language.

We then generate textual questions and answers based on the ground-truth annotations for each task with textual LLMs. For the first 8 tasks, we use the input speech as a prompt to clone a synthesized question speech and randomly concatenate the question before or after the original speech. For the remaining 3 tasks, we further mix these audios with synthesized speeches before question concatenation, creating more challenging test samples.

We also establish an automatic evaluation protocol for StepEval-Audio-Paralinguistic, which initially transcribes model outputs into text using ASR, followed by automatic judgment with a textual LLM. More information, along with the complete StepEval-Audio-Paralinguistic test set and evaluation code, is available at \url{https://github.com/stepfun-ai/Step-Audio2} to foster further research on paralinguistic information understanding.

We evaluate GPT-4o Audio, Kimi-Audio, Qwen-Omni, Step-Audio-AQAA, and Step-Audio 2 using the StepEval-Audio-Paralinguistic benchmark, with results presented in Table~\ref{tab:paralinguistic}. The experimental results highlight the comprehensive capabilities of Step-Audio 2 in understanding various paralinguistic information, achieving an average accuracy of 83.09, which is a significant improvement over other baseline models.

\begin{table}[htb]
\caption{Comparison between GPT-4o Audio, Kimi-Audio, Qwen-Omni, Step-Audio-AQAA and Step-Audio 2 on StepEval-Audio-Paralinguistic.}
\label{tab:paralinguistic}
\centering
\begin{tabular}{lcccccccccccc}
\toprule
\textbf{Model} &\textbf{Avg.} &\textbf{Gender} & \textbf{Age} & \textbf{Timbre} & \textbf{Scenario} & \textbf{Event}  \\
\midrule
\textbf{GPT-4o Audio} & 43.45 & 18 & 42 & 34 & 22 & 14 \\
\textbf{Kimi-Audio} & 49.64 & 94 & 50 & 10 & 30 & 48 & \\
\textbf{Qwen-Omni} & 44.18 & 40 & 50 & 16 & 28 & 42  \\
\textbf{Step-Audio-AQAA} & 36.91 & 70 & 66 & 18 & 14 & 14  \\
\textbf{Step-Audio 2} & \textbf{83.09} & \textbf{100} & \textbf{96} & \textbf{82} & \textbf{78} & \textbf{60} \\
\midrule
\textbf{Model} &\textbf{Emotion} & \textbf{Pitch} & \textbf{Rhythm} & \textbf{Speed} & \textbf{Style} & \textbf{Vocal} \\
\midrule
\textbf{GPT-4o Audio} & 82 & 40 & 60 & 58 & 64 & 44  \\
\textbf{Kimi-Audio} & 66 & 56 & 40 & 44 & 54 & 54 \\
\textbf{Qwen-Omni} & 76 & 32 & 54 & 50 & 50 & 48 \\
\textbf{Step-Audio-AQAA} & 40 & 38 & 48 & 54 & 44 & 0  \\
\textbf{Step-Audio 2} & \textbf{86} & \textbf{82} & \textbf{86} & \textbf{88} & \textbf{88} & \textbf{68} \\
\bottomrule
\end{tabular}
\end{table}

\subsection{Audio understanding}

We then assess Step-Audio 2's general audio comprehension across sound, speech, and music using the latest version of the MMAU  benchmark~\cite{sakshi2024mmau}\footnote{MMAU v05.15.25 test-mini}.

As baselines, we employ Audio Flamingo 3, Gemini 2.5 Pro, GPT-4o Audio, Kimi-Audio, Omni-R1~\cite{rouditchenko2025omni}, Qwen2.5-Omni, and Step-Audio-AQAA. We obtain the reported results for Audio Flamingo 3, Omni-R1, and Qwen2.5-Omni from their original papers. The results of Gemini 2.5 Pro are obtained from the official website of MMAU.
And we re-evaluate GPT-4o Audio, Kimi-Audio and Step-Audio-AQAA due to the recent update of the MMAU benchmark.

The results are summarized in Table~\ref{tab:MMAU}. Step-Audio 2 achieves the highest average score of 78.0, followed by Omni-R1 and Audio Flamingo 3, both of which are specialized approaches in audio understanding.
Specifically, Step-Audio 2 yields the best results in sound and speech tracks and on par results with the best in music track, demonstrating its versatility and robustness across different audio domains.

\begin{table}[htb]
\caption{Comparison between Audio Flamingo 3, Gemini 2.5 Pro, GPT-4o Audio, Kimi-Audio, Omni-R1, Qwen2.5-Omni, Step-Audio-AQAA and Step-Audio 2 on MMAU.}
\label{tab:MMAU}
\centering
\begin{tabular}{lcccc}
\toprule
\textbf{Model}& \textbf{Avg.} & \textbf{Sound} & \textbf{Speech} & \textbf{Music}  \\
\midrule
\textbf{Audio Flamingo 3} &73.1 & 76.9& 66.1& \textbf{73.9}\\
\textbf{Gemini 2.5 Pro} & 71.6  & 75.1& 71.5 &68.3\\
\textbf{GPT-4o Audio} & 58.1 & 58.0& 64.6 &51.8\\
\textbf{Kimi-Audio} & 69.6& 79.0& 65.5&64.4\\
\textbf{Omni-R1}&77.0 & 81.7& 76.0&73.4\\
\textbf{Qwen2.5-Omni} &71.5 & 78.1& 70.6&65.9\\
\textbf{Step-Audio-AQAA} & 49.7 &  50.5&  51.4& 47.3\\
\textbf{Step-Audio 2} & \textbf{78.0} & \textbf{83.5} & \textbf{76.9} & 73.7\\
\bottomrule
\end{tabular}
\end{table}

\subsection{Speech translation}
We evaluate the model's bidirectional Chinese-English speech translation capabilities using two benchmarks: speech-to-text translation (S2TT) on CoVoST 2~\cite{wang2020covost2massivelymultilingual} and speech-to-speech translation (S2ST) on CVSS~\cite{jia2022cvsscorpusmassivelymultilingual}.
Additionally, we use the reported results of Qwen2.5-Omni for CoVoST~2, while for CVSS, we employ Qwen-Omni as a baseline.
Kimi-Audio is excluded because it consistently ignores prompts and performs ASR instead of translation. Using BLEU as the evaluation metric, the results in Table~\ref{tab:S2ST} demonstrate that Step-Audio 2 achieves superior performance in Chinese-English bidirectional translations, obtaining the highest average score on both the CoVoST~2 and CVSS test sets.

\begin{table}[htb]
\caption{Comparison of BLEU scores between GPT-4o Audio, Qwen2.5-Omni, Qwen-Omni, Step-Audio-AQAA and Step-Audio 2 on speech-to-text and speech-to-speech translation.}
\label{tab:S2ST}
\centering
\begin{tabular}{lccc}
\toprule
\multirow{2}{*}{\textbf{Model}}&\multicolumn{3}{c}{\textbf{CoVoST 2 (Speech-to-text translation)}}\\
\cmidrule(lr){2-4}
& \textbf{Avg.} & \textbf{English-to-Chinese}& \textbf{Chinese-to-English}\\
\midrule
\textbf{GPT-4o Audio}       &29.61& 40.20 & 19.01 \\
\textbf{Qwen2.5-Omni}       &35.40& 41.40 & 29.40 \\
\textbf{Step-Audio-AQAA}    &28.57& 37.71 & 19.43 \\
\textbf{Step-Audio 2}       &\textbf{39.26}& \textbf{49.01} & \textbf{29.51} \\
\midrule
\multirow{2}{*}{\textbf{Model}}&\multicolumn{3}{c}{\textbf{CVSS (Speech-to-speech translation)}}\\
\cmidrule(lr){2-4}
&\textbf{Avg.} & \textbf{English-to-Chinese} & \textbf{Chinese-to-English} \\
\midrule
\textbf{GPT-4o Audio}       &23.68& 20.07 & \textbf{27.29} \\
\textbf{Qwen-Omni}          &15.35& 8.04  & 22.66 \\
\textbf{Step-Audio-AQAA}    &27.36& 30.74 & 23.98 \\
\textbf{Step-Audio 2}       &\textbf{30.87}& \textbf{34.83} & 26.92\\
\bottomrule
\end{tabular}
\end{table}

\subsection{Tool calling}

To address the gap in the availability of suitable test sets for tool calling in speech conversations, we introduce StepEval-Audio-Toolcall, a test set that evaluates the model's ability in tool invocation, selection and parameter extraction under Chinese speech conversations.

We employ a textual LLM to generate 200 multi-turn dialogue scripts for each kind of tool.
Each script contains 3-6 turns of inputs and outputs, in which
previous turns may or may not include tool calling statements,
but the final input must contain a calling intention to a specific external tool.
We then balance the samples with an equal number of negative samples for each kind of tools, 
in which the final speech input either has no tool calling intention or intention to call on other kinds of tools. 
Subsequently, we synthesize these scripts into speeches with our conversation synthesis pipeline.
And we propose an automatic evaluation protocol to employ Qwen3-32B to automatically examine the output and tool calling statements.  
We release StepEval-Audio-Toolcall including the original scripts, synthesized speech conversations and the corresponding evaluation script in \url{https://github.com/stepfun-ai/Step-Audio2}.

Despite that there is no other LALM that provides custom tool calling,
we employ Qwen3-32B as a baseline to illustrate how Step-Audio 2 manages external tools in comparison to textual LLMs.
As shown in Table~\ref{tab:toolcall}, 
Step-Audio 2 achieves on par with tool calling accuracy with textual LLMs even with speech input.
Notably, Step-Audio 2 significantly
outperforms Qwen3-32B in accurately calling our innovative audio search tool,
highlighting its specialty as a multi-modal LLM than textual LLMs.

\begin{table}[htb]
    \centering
    \caption{Comparison between Step-Audio 2 and Qwen3-32B on StepEval-Audio-Toolcall. $^\dagger$Qwen3-32B is evaluated with text inputs. $^\ddagger$Date and time tools have no parameter.}
    \begin{tabular}{llccccc}
    \toprule
        \textbf{Model} & \textbf{Objective} & \textbf{Metric} 
        & \textbf{Audio search} & \textbf{Date \& Time$^\ddagger$} & \textbf{Weather} & \textbf{Web search}  \\
        \midrule
        & \textbf{Trigger} & \textbf{Precision / Recall}& 67.5 / 98.5 & 98.4 / 100.0 & 90.1 / 100.0 & 86.8 / 98.5 \\
        \textbf{Qwen3-32B}$^\dagger$  & \textbf{Type} & \textbf{Accuracy} & 100.0 & 100.0 & 98.5 & 98.5 \\
        & \textbf{Parameter} & \textbf{Accuracy} & 100.0 & N/A & 100.0 & 100.0\\
        \cmidrule(lr){2-7}
        & \textbf{Trigger} &\textbf{Precision / Recall} & 86.8 / 99.5 & 96.9 / 98.4 & 92.2 / 100.0 & 88.4 / 95.5 \\
        \textbf{Step-Audio 2}  & \textbf{Type} & \textbf{Accuracy}  & 100.0 & 100.0 & 90.5 & 98.4 \\
        & \textbf{Parameter} & \textbf{Accuracy}  & 100.0& N/A& 100.0& 100.0\\
        \bottomrule
    \end{tabular}
    \label{tab:toolcall}
\end{table}

\subsection{Speech-to-speech conversation}
We finally employ URO-Bench \cite{yan2025urobenchcomprehensivebenchmarkendtoend}
to evaluate Step-Audio 2 and other open-source and commercial LALMs, including GPT-4o Audio, Kimi-Audio, Qwen-Omni, and Step-Audio-AQAA.
URO-Bench consists of multiple datasets on two difficulty tracks, evaluating the model's understanding, reasoning and oral conversation abilities, such as ASR, instruction following, commonsense knowledge, mathematics, and speech naturalness, emotion and speaking styles expressions.
We follow the ASR-mediated procedure in URO-Bench for evaluation, employing Whisper for ASR and GPT-4o-mini for automatic judging.

As demonstrated in Table~\ref{tab:uro}, Step-Audio 2 significantly outperforms existing large audio language models, including GPT-4o Audio, in Chinese speech-to-speech conversation scenarios, achieving the highest average scores of 83.32 on the basic track and 68.25 on the pro track. In English speech-to-speech conversations, while Step-Audio 2 is slightly outperformed by GPT-4o Audio, it provides very competitive results and exceeds the other approaches.

\begin{table}[ht]
\centering
\caption{Comparison between GPT-4o Audio, Kimi-Audio, Qwen-Omni, Step-Audio-AQAA and Step-Audio 2 on the URO-Bench. U. R. O. stands for understanding, reasoning, and oral conversation, respectively.}
\begin{tabular}{lccccccccc}
\toprule
\multirow{2}{*}{\textbf{Model}} & \multirow{2}{*}{\textbf{Language}} & \multicolumn{4}{c}{\textbf{Basic}}& \multicolumn{4}{c}{\textbf{Pro}}\\
\cmidrule(lr){3-6} \cmidrule(lr){7-10}
&&\textbf{Avg.} & \textbf{U.} & \textbf{R.} & \textbf{O.} & \textbf{Avg.} & \textbf{U.} & \textbf{R.} & \textbf{O.} \\
\midrule
\textbf{GPT-4o Audio}&  \multirow{6}{*}{\textbf{Chinese}}&78.59&89.40 &65.48 &85.24 &67.10&70.60 &57.22 & 70.20 \\
\textbf{Kimi-Audio} &                                    &73.59&79.34 &64.66 &79.75 &66.07&60.44 &59.29 & \textbf{76.21} \\
\textbf{Qwen-Omni} &                                     &68.98&59.66 &69.74 &77.27 &59.11&59.01 &59.82 & 58.74 \\
\textbf{Step-Audio-AQAA} &                               &74.71&87.61 &59.63 &81.93 &65.61&74.76  &47.29  & 68.97 \\
\textbf{Step-Audio 2} &                                  &\textbf{83.32}&\textbf{91.05} &\textbf{75.45} & \textbf{86.08} &\textbf{68.25} & \textbf{74.78} &\textbf{63.18} & 65.10  \\
\midrule
\textbf{GPT-4o Audio}&  \multirow{5}{*}{\textbf{English}}&\textbf{84.54}&90.18 &75.90 &\textbf{90.41} &\textbf{67.51}&60.65 &64.36 &\textbf{78.46} \\
\textbf{Kimi-Audio} &                                    &60.04&83.36 &42.31 &60.36 &49.79&50.32 &40.59 &56.04 \\
\textbf{Qwen-Omni} &                                     &70.58&66.29 &69.62 &76.16 &50.99&44.51 &63.88 &49.41 \\
\textbf{Step-Audio-AQAA} &                               &71.11&90.15 &56.12 &72.06 &52.01&44.25 &54.54 &59.81 \\
\textbf{Step-Audio 2} &                                  &83.90&\textbf{92.72} &\textbf{76.51} &84.92 &66.07 &\textbf{64.86} &\textbf{67.75}&66.33 \\
\bottomrule
\end{tabular}
\label{tab:uro}
\end{table}

\section{Conclusion}
We introduce Step-Audio~2, an end-to-end large audio language model designed for enterprise speech and audio understanding, as well as intelligent speech interaction. Step-Audio~2 leverages a latent audio encoder and reinforcement learning to enhance its speech and audio comprehension capabilities. Furthermore, by integrating the generation of discrete audio tokens into language modeling, Step-Audio~2 achieves genuine end-to-end speech interaction and improves its responsiveness to paralinguistic information, such as speaking styles and emotions. Step-Audio~2 is also capable of utilizing external tools including web search and audio search for multi-modal RAG. Trained on 8 million hours of speeches and audios, Step-Audio~2 demonstrates state-of-the-art performance across various tasks, including ASR, audio understanding, speech translation, and general speech conversation, outperforming both open-source and commercial solutions.

\printbibliography[title={References}]

\newpage
\appendix

\section*{Appendix}
\section{Contributors}
The contributors are list in alphabet order. 
\subsection{Core contributors}
\paragraph{Model}
Boyong Wu,
Chao Yan,
Chen Hu,
Cheng Yi,
Chengli Feng,
Fei Tian,
Feiyu Shen, 
Gang Yu,
Haoyang Zhang,
Jingbei Li,
Mingrui Chen, 
Peng Liu,
Wang You,
Xiangyu (Tony) Zhang,
Xingyuan Li,
Xuerui Yang,
Yayue Deng,
Yechang Huang,
Yuxin Li,
Yuxin Zhang,
Zhao You

\paragraph{Infrastructure}
Brian Li, %
Changyi Wan,
Hanpeng Hu,
Jiangjie Zhen,
Siyu Chen,
Song Yuan,
Xuelin Zhang,
Yimin Jiang,
Yu Zhou,
Yuxiang Yang

\paragraph{Data and evaluation}
Bingxin Li, %
Buyun Ma, %
Changhe Song, %
Dongqing Pang, %
Guoqiang Hu, %
Haiyang Sun, %
Kang An, %
Na Wang, %
Shuli Gao, %
Wei Ji,  %
Wen Li, %
Wen Sun, %
Xuan Wen, %
Yong Ren, %
Yuankai Ma, %
Yufan Lu %

\subsection{Contributors}
Bin Wang, %
Bo Li, %
Changxin Miao, %
Che Liu, %
Chen Xu, %
Dapeng Shi, %
Dingyuan Hu, %
Donghang Wu, %
Enle Liu, %
Guanzhe Huang, %
Gulin Yan, %
Han Zhang,
Hao Nie, %
Haonan Jia, %
Hongyu Zhou, %
Jianjian Sun, %
Jiaoren Wu, %
Jie Wu, %
Jie Yang, %
Jin Yang, %
Junzhe Lin, %
Kaixiang Li, %
Lei Yang,
Liying Shi, %
Li Zhou, %
Longlong Gu, %
Ming Li, %
Mingliang Li, %
Mingxiao Li, %
Nan Wu, %
Qi Han,
Qinyuan Tan, %
Shaoliang Pang, %
Shengjie Fan, %
Siqi Liu, %
Tiancheng Cao, %
Wanying Lu, %
Wenqing He, %
Wuxun Xie, %
Xu Zhao, %
Xueqi Li, %
Yanbo Yu, %
Yang Yang, %
Yi Liu, %
Yifan Lu, %
Yilei Wang, %
Yuanhao Ding, %
Yuanwei Liang, %
Yuanwei Lu, %
Yuchu Luo, %
Yuhe Yin, %
Yumeng Zhan, %
Yuxiang Zhang, %
Zidong Yang, %
Zixin Zhang %

\subsection{Sponsors}
Binxing Jiao,
Daxin Jiang,
Heung-Yeung Shum,
Jiansheng Chen,
Jing Li,
Xiangyu Zhang,
Yibo Zhu

\subsection{External contributors}
\paragraph{Nanyang Technological University (NTU), Singapore}
Eng Siong Chng,
Hexin Liu
\newpage

\section{Introduction and evaluation results of Step-Audio 2 mini}

We are pleased to release Step-Audio 2 mini, a special open-source version of Step-Audio 2, available at \url{https://github.com/stepfun-ai/Step-Audio2}.
Step-Audio 2 mini employs the encoder from Qwen2-Audio as its audio encoder and is initialized with Qwen2.5-7B.
Step-Audio 2 mini is trained on the same dataset as Step-Audio 2, but it is limited to utilize only the web search tool.

Step-Audio 2 mini is a more developer-friendly variant of Step-Audio 2, with a parameter count for fair comparisons with open-source models including Qwen-Omni and Kimi-Audio. Evaluation results\footnote{Evaluation results are obtained with our vLLM backend and may differ from the results with transformers backend.} demonstrate that Step-Audio 2 mini delivers on par results with Step-Audio 2, exceeding most open-source and commercial models such as GPT-4o Audio.

\subsection{Automatic speech recognition}
\begin{table}[htb]
\centering
\caption{Comparison between Doubao LLM ASR, GPT-4o Transcribe, Kimi-Audio, Qwen-Omni, Step-Audio 2 and Step-Audio 2 mini, on character (for Chinese, Cantonese and Japanese) and word (for Arabian and English) error rates among multiple ASR test sets. N/A indicates that the language is not supported.}
\label{tab:asr-7b}
\begin{tabular}{lccccccc}
\toprule
\textbf{Category} & \textbf{Test set} & \textbf{\makecell{Doubao\\LLM ASR}} & \textbf{\makecell{GPT-4o \\Transcribe}} & \textbf{\makecell{Kimi-\\Audio}} & \textbf{\makecell{Qwen-\\Omni}} & \textbf{\makecell{Step-\\Audio 2}} & \textbf{\makecell{Step-Audio\\2 mini}}\\
\midrule
\multirow{5}{*}{\textbf{English}} 
& Common Voice & 9.20 & 9.30 & 7.83 & 8.33 & \textbf{5.95} & 6.76\\
& FLEURS English & 7.22 & \textbf{2.71} & 4.47 & 5.05 & 3.03 & 3.05\\
& LibriSpeech clean & 2.92 & 1.75 & 1.49 & 2.93 & \textbf{1.17} & 1.33\\
& LibriSpeech other & 5.32 & 4.23 & 2.91 & 5.07 & \textbf{2.42} & 2.86\\
\cmidrule(lr){2-8}
& \textbf{Average} & 6.17 & 4.50 & 4.18 & 5.35 & \textbf{3.14} & 3.50\\
\cmidrule(r){1-8}
\multirow{7}{*}{\textbf{Chinese}}
& AISHELL & 0.98 & 3.52 & 0.64 & 1.17 & \textbf{0.63} & 0.78\\
& AISHELL-2 & 3.10 & 4.26 & 2.67 & 2.40 & \textbf{2.10} & 2.16\\
& FLEURS Chinese & 2.92 & 2.62 & 2.91 & 7.01 & 2.68 & \textbf{2.53}\\
& KeSpeech phase1& 6.48 & 26.80 & 5.11 & 6.45 & \textbf{3.63} & 3.97\\
& WenetSpeech meeting & 4.90 & 31.40 & 5.21 & 6.61 & \textbf{4.75} & 4.87\\
& WenetSpeech net & \textbf{4.46} & 15.71 & 5.93 & 5.24 & 4.67 & 4.82\\
\cmidrule(lr){2-8}
& \textbf{Average} & 3.81 & 14.05 & 3.75 & 4.81 & \textbf{3.08} & 3.19\\
\cmidrule(r){1-8}
& FLEURS Arabian & N/A & \textbf{11.72} & N/A & 25.13 & 14.22 & 16.46\\
\textbf{Multilingual}& Common Voice yue & 9.20 & 11.10 & 38.90 & \textbf{7.89} & 7.90 & 8.32\\
& FLEURS Japanese & N/A & 3.27 & N/A & 10.49 & \textbf{3.18} & 4.67\\
\cmidrule(r){1-8}
 & Anhui accent & \textbf{8.83} & 50.55 & 22.17 & 18.73 & 10.61 & 11.65\\
 & Guangdong accent & 4.99 & 7.83 & \textbf{3.76} & 4.03 & 3.81 & 4.44\\
 & Guangxi accent & 3.37 & 7.09 & 4.29 & \textbf{3.35} & 4.11 & 3.51\\
\textbf{In-house} & Shanxi accent & 20.26 & 55.03 & 34.71 & 25.95 & \textbf{12.44} & 15.60\\
 & Sichuan dialect & \textbf{3.01} & 32.85 & 5.26 & 5.61 & 4.35 & 4.57\\
& Shanghai dialect & 47.49 & 89.58 & 82.90 & 58.74 & \textbf{17.77} & 19.30\\
\cmidrule(lr){2-8}
& \textbf{Average} & 14.66 & 40.49 & 25.52 & 19.40 & \textbf{8.85} & 9.85\\
\bottomrule
\end{tabular}
\end{table}

\newpage
\subsection{Paralinguistic information understanding}

\begin{table}[h]
\caption{Comparison between GPT-4o Audio, Kimi-Audio, Qwen-Omni, Step-Audio-AQAA, Step-Audio 2 and Step-Audio 2 mini on StepEval-Audio-Paralinguistic.}
\label{tab:paralinguistic-7b}
\centering
\begin{tabular}{lcccccccccccc}
\toprule
\textbf{Model} &\textbf{Avg.} &\textbf{Gender} & \textbf{Age} & \textbf{Timbre} & \textbf{Scenario} & \textbf{Event}  \\
\midrule
\textbf{GPT-4o Audio} & 43.45 & 18 & 42 & 34 & 22 & 14 \\
\textbf{Kimi-Audio} & 49.64 & 94 & 50 & 10 & 30 & 48 \\
\textbf{Qwen-Omni} & 44.18 & 40 & 50 & 16 & 28 & 42  \\
\textbf{Step-Audio-AQAA} & 36.91 & 70 & 66 & 18 & 14 & 14  \\
\textbf{Step-Audio 2} & \textbf{83.09} & \textbf{100} & \textbf{96} & \textbf{82} & \textbf{78} & \textbf{60} \\
\textbf{Step-Audio 2 mini} & 80.00 & \textbf{100} & 94 & 80 & \textbf{78} & \textbf{60} \\
\midrule
\textbf{Model} &\textbf{Emotion} & \textbf{Pitch} & \textbf{Rhythm} & \textbf{Speed} & \textbf{Style} & \textbf{Vocal} \\
\midrule
\textbf{GPT-4o Audio} & 82 & 40 & 60 & 58 & 64 & 44  \\
\textbf{Kimi-Audio} & 66 & 56 & 40 & 44 & 54 & 54 \\
\textbf{Qwen-Omni} & 76 & 32 & 54 & 50 & 50 & 48 \\
\textbf{Step-Audio-AQAA} & 40 & 38 & 48 & 54 & 44 & 0  \\
\textbf{Step-Audio 2} & \textbf{86} & \textbf{82} & \textbf{86} & \textbf{88} & \textbf{88} & 68 \\
\textbf{Step-Audio 2 mini} & 82 & \textbf{82} & 68 & 74 & 86 & \textbf{76} \\
\bottomrule
\end{tabular}
\end{table}

\subsection{Audio understanding}
\begin{table}[h]
\caption{Comparison between Audio Flamingo 3, Gemini 2.5 Pro, GPT-4o Audio, Kimi-Audio, Omni-R1, Qwen2.5-Omni, Step-Audio-AQAA, Step-Audio 2 and Step-Audio 2 mini on MMAU.}
\label{tab:MMAU-7b}
\centering
\begin{tabular}{lcccc}
\toprule
\textbf{Model}& \textbf{Avg.} & \textbf{Sound} & \textbf{Speech} & \textbf{Music}  \\
\midrule
\textbf{Audio Flamingo 3} &73.1 & 76.9& 66.1& \textbf{73.9}\\
\textbf{Gemini 2.5 Pro} & 71.6  & 75.1& 71.5 &68.3\\
\textbf{GPT-4o Audio} & 58.1 & 58.0& 64.6 &51.8\\
\textbf{Kimi-Audio} & 69.6& 79.0& 65.5&64.4\\
\textbf{Omni-R1}&77.0 & 81.7& 76.0&73.4\\
\textbf{Qwen2.5-Omni} &71.5 & 78.1& 70.6&65.9\\
\textbf{Step-Audio-AQAA} & 49.7 &  50.5&  51.4& 47.3\\
\textbf{Step-Audio 2} & \textbf{78.0} & \textbf{83.5} & \textbf{76.9} & 73.7\\
\textbf{Step-Audio 2 mini} & 73.2 & 76.6 & 71.5 & 71.6\\
\bottomrule
\end{tabular}
\end{table}
\newpage

\subsection{Speech translation}
\begin{table}[h]
\caption{Comparison of BLEU scores between GPT-4o Audio, Qwen2.5-Omni, Qwen-Omni, Step-Audio-AQAA, Step-Audio 2 and Step-Audio 2 mini on speech-to-text and speech-to-speech translation.}
\label{tab:S2ST-7b}
\centering
\begin{tabular}{lccc}
\toprule
\multirow{2}{*}{\textbf{Model}}&\multicolumn{3}{c}{\textbf{CoVoST 2 (Speech-to-text translation)}}\\
\cmidrule(lr){2-4}
& \textbf{Avg.} & \textbf{English-to-Chinese}& \textbf{Chinese-to-English}\\
\midrule
\textbf{GPT-4o Audio}       &29.61& 40.20 & 19.01 \\
\textbf{Qwen2.5-Omni}       &35.40& 41.40 & 29.40 \\
\textbf{Step-Audio-AQAA}    &28.57& 37.71 & 19.43 \\
\textbf{Step-Audio 2} &39.26& 49.01 & \textbf{29.51} \\
\textbf{Step-Audio 2 mini}  &\textbf{39.29}& \textbf{49.12} & 29.47 \\
\midrule
\multirow{2}{*}{\textbf{Model}}&\multicolumn{3}{c}{\textbf{CVSS (Speech-to-speech translation)}}\\
\cmidrule(lr){2-4}
&\textbf{Avg.} & \textbf{English-to-Chinese} & \textbf{Chinese-to-English} \\
\midrule
\textbf{GPT-4o Audio}       &23.68& 20.07 & \textbf{27.29} \\
\textbf{Qwen-Omni}          &15.35& 8.04  & 22.66 \\
\textbf{Step-Audio-AQAA}    &27.36& 30.74 & 23.98 \\
\textbf{Step-Audio 2} &\textbf{30.87}& \textbf{34.83} & 26.92 \\
\textbf{Step-Audio 2 mini}  &29.08& 32.81 & 25.35 \\
\bottomrule
\end{tabular}
\end{table}

\subsection{Speech-to-speech conversation}
\begin{table}[h]
\centering
\caption{Comparison between GPT-4o Audio, Kimi-Audio, Qwen-Omni, Step-Audio-AQAA, Step-Audio 2 and Step-Audio 2 mini on the URO-Bench. U. R. O. stands for understanding, reasoning, and oral conversation, respectively.}
\begin{tabular}{lccccccccc}
\toprule
\multirow{2}{*}{\textbf{Model}} & \multirow{2}{*}{\textbf{Language}} & \multicolumn{4}{c}{\textbf{Basic}}& \multicolumn{4}{c}{\textbf{Pro}}\\
\cmidrule(lr){3-6} \cmidrule(lr){7-10}
&&\textbf{Avg.} & \textbf{U.} & \textbf{R.} & \textbf{O.} & \textbf{Avg.} & \textbf{U.} & \textbf{R.} & \textbf{O.} \\
\midrule
\textbf{GPT-4o Audio}&  \multirow{6}{*}{\textbf{Chinese}}&78.59&89.40 &65.48 &85.24 &67.10&70.60 &57.22 & 70.20 \\
\textbf{Kimi-Audio} &                                    &73.59&79.34 &64.66 &79.75 &66.07&60.44 &59.29 & \textbf{76.21} \\
\textbf{Qwen-Omni} &                                     &68.98&59.66 &69.74 &77.27 &59.11&59.01 &59.82 & 58.74 \\
\textbf{Step-Audio-AQAA} &                               &74.71&87.61 &59.63 &81.93 &65.61&74.76  &47.29  & 68.97 \\
\textbf{Step-Audio 2} &                                  &\textbf{83.32}&\textbf{91.05} &\textbf{75.45} & \textbf{86.08} &68.25 &74.78 &\textbf{63.18} & 65.10  \\
\textbf{Step-Audio 2 mini} &                             &77.81&89.19 &64.53 &84.12 &\textbf{69.57}&\textbf{76.84} &58.90 & 69.42 \\
\midrule
\textbf{GPT-4o Audio}&  \multirow{6}{*}{\textbf{English}}&\textbf{84.54}&90.18 &75.90 &\textbf{90.41} &\textbf{67.51}&60.65 &64.36 &\textbf{78.46} \\
\textbf{Kimi-Audio} &                                    &60.04&83.36 &42.31 &60.36 &49.79&50.32 &40.59 &56.04 \\
\textbf{Qwen-Omni} &                                     &70.58&66.29 &69.62 &76.16 &50.99&44.51 &63.88 &49.41 \\
\textbf{Step-Audio-AQAA} &                               &71.11&90.15 &56.12 &72.06 &52.01&44.25 &54.54 &59.81 \\
\textbf{Step-Audio 2} &                                  &83.90&\textbf{92.72} &\textbf{76.51} &84.92 &66.07 &\textbf{64.86} &\textbf{67.75}&66.33 \\
\textbf{Step-Audio 2 mini} &                             &74.36&90.07 &60.12 &77.65 &61.25&58.79 &61.94 & 63.80 \\
\bottomrule
\end{tabular}
\label{tab:uro-7b}
\end{table}

\end{document}